\documentclass[10pt,journal]{IEEEtran} % <--- FOR ARXIV

\usepackage[utf8]{inputenc}
\usepackage{booktabs}
\usepackage{multirow}
\usepackage{makecell}
\usepackage{graphicx}
\usepackage{tabularx}
\usepackage{amsmath}
\usepackage{amssymb}
\usepackage{float}
\ifCLASSOPTIONcompsoc
  \usepackage[nocompress]{cite}
\else
  \usepackage{cite}
\fi
\usepackage{amsmath}
\usepackage{array}
\ifCLASSOPTIONcompsoc
  \usepackage[caption=false,font=footnotesize,labelfont=sf,textfont=sf]{subfig}
\else
  \usepackage[caption=false,font=footnotesize]{subfig}
\fi
\usepackage{fixltx2e}
\usepackage{url}
\begin{document}
%
% paper title
% Titles are generally capitalized except for words such as a, an, and, as,
% at, but, by, for, in, nor, of, on, or, the, to and up, which are usually
% not capitalized unless they are the first or last word of the title.
% Linebreaks \\ can be used within to get better formatting as desired.
% Do not put math or special symbols in the title.
\title{Unsupervised Multi-Modal Representation Learning for Affective Computing with Multi-Corpus Wearable Data}
%
%
% author names and IEEE memberships
% note positions of commas and nonbreaking spaces ( ~ ) LaTeX will not break
% a structure at a ~ so this keeps an author's name from being broken across
% two lines.
% use \thanks{} to gain access to the first footnote area
% a separate \thanks must be used for each paragraph as LaTeX2e's \thanks
% was not built to handle multiple paragraphs
%
%
%\IEEEcompsocitemizethanks is a special \thanks that produces the bulleted
% lists the Computer Society journals use for "first footnote" author
% affiliations. Use \IEEEcompsocthanksitem which works much like \item
% for each affiliation group. When not in compsoc mode,
% \IEEEcompsocitemizethanks becomes like \thanks and
% \IEEEcompsocthanksitem becomes a line break with idention. This
% facilitates dual compilation, although admittedly the differences in the
% desired content of \author between the different types of papers makes a
% one-size-fits-all approach a daunting prospect. For instance, compsoc 
% journal papers have the author affiliations above the "Manuscript
% received ..."  text while in non-compsoc journals this is reversed. Sigh.

\author{Kyle~Ross, Paul~Hungler, and Ali~Etemad% <-this % stops a space
\IEEEcompsocitemizethanks{\IEEEcompsocthanksitem K. Ross and A. Etemad are with the Department of Electrical and Computer Engineering, Queen's University, Kingston,
ON, K7L 3N6.\protect\\
% note need leading \protect in front of \\ to get a newline within \thanks as
% \\ is fragile and will error, could use \hfil\break instead.
\IEEEcompsocthanksitem P. Hungler is with the Faculty of Engineering and Applied Science, Queen's University.}% <-this % stops an unwanted space
\thanks{}}

\IEEEtitleabstractindextext{%
\begin{abstract}
With recent developments in smart technologies, there has been a growing focus on the use of artificial intelligence and machine learning for affective computing to further enhance the user experience through emotion recognition. Typically, machine learning models used for affective computing are trained using manually extracted features from biological signals. Such features may not generalize well for large datasets and may be sub-optimal in capturing the information from the raw input data. One approach to address this issue is to use fully supervised deep learning methods to learn latent representations of the biosignals. However, this method requires human supervision to label the data, which may be unavailable or difficult to obtain. In this work we propose an unsupervised framework reduce the reliance on human supervision. The proposed framework utilizes two stacked convolutional autoencoders to learn latent representations from wearable electrocardiogram (ECG) and electrodermal activity (EDA) signals. These representations are utilized within a random forest model for binary arousal classification. This approach reduces human supervision and enables the aggregation of datasets allowing for higher generalizability. To validate this framework, an aggregated dataset comprised of the AMIGOS, ASCERTAIN, CLEAS, and MAHNOB-HCI datasets is created. The results of our proposed method are compared with using convolutional neural networks, as well as methods that employ manual extraction of hand-crafted features. The methodology used for fusing the two modalities is also investigated. Lastly, we show that our method outperforms current state-of-the-art results that have performed arousal detection on the same datasets using ECG and EDA biosignals. The results show the wide-spread applicability for stacked convolutional autoencoders to be used with machine learning for affective computing.
\end{abstract}
}
% Note that keywords are not normally used for peerreview papers.
%\begin{IEEEkeywords}
%Computer Society, IEEE, IEEEtran, journal, \LaTeX, paper, template.
%\end{IEEEkeywords}}

\maketitle

\IEEEdisplaynontitleabstractindextext

\IEEEpeerreviewmaketitle

% \IEEEraisesectionheading{\section{Introduction}\label{sec:introduction}}

\section{Introduction}\label{sec:introduction}

\IEEEPARstart{S}{mart} technologies are quickly becoming ubiquitous in our everyday lives. These technologies aim to understand, analyze, and interact with users seamlessly, providing a user-centered experience. Affective computing is one area of research aiming to create smart technologies that can better understand and ultimately react to their user. Affective computing is the notion of detecting, modeling, and reacting to the user's affective states by a computer \cite{picard_2000}. The integration of affective computing into devices can make users perceive the computers as more intelligent, effectively making the smart technology ``smarter'' \cite{Picard2001}.
% You must have at least 2 lines in the paragraph with the drop letter
% (should never be an issue)

With the increasing pervasiveness of smart technologies, research into determining the user's affective state to facilitate affective computing is of growing importance. Affective states can be broken down into two dimensions, valence and arousal \cite{mehrabian1974approach}. Both valence and arousal have been shown to be closely linked with activity in the autonomic nervous system \cite{GOMEZ201651}. Biosignals are one source through which changes in the autonomic nervous system can be observed, allowing for changes in valence and arousal to be elucidated \cite{Greene}. Accordingly, machine learning approaches utilizing biosignals have been widely used for the classification of valence and arousal.

To collect the biosignals for affective computing through machine learning, wearable sensors have been widely used. While original wearable devices were bulky and cumbersome to use, with recent advancements, they have become lightweight, affordable, and unobtrusive \cite{Sperlich1240}. In fact, wearables have become small enough to be incorporated into jewelry and clothing so that they can be used in everyday life \cite{PicardHealey}. It is estimated that by 2022 the market for wearables will double to be worth 27 billion dollars \cite{Russey_wearables}. This will result in 233 million total sales of wearable devices \cite{lamkin_2018}.

Wearable devices that collect Electrocardiogram (ECG) \cite{wan2009, Agrafioti}, Electrodermal Activity (EDA) \cite{feng2018, hassan2019}, Electroencephalogram (EEG) \cite{rozgic2013, murugappan2010}, Electrooculogram (EOG) \cite{Anderson2017, Shin2018}, and Electromyography (EMG) \cite{murugappan2011, yang2011} have all been used for a variety of applications including affective computing. Among these signals, ECG and EDA have been shown to be the most closely correlated to arousal \cite{Healey}.

Typically, machine learning models used for affective computing are trained using time domain and frequency domain features that are hand-crafted and manually extracted from biosignals \cite{ayata2016, ayata2017, wang2011, Petrantonakis2012}. However, these features may not generalize well between multiple datasets as the biosignals can be vastly different in terms of the quality of the data, the placement of sensors on the body, and the data collection protocol as a whole. Another approach for extracting biosignal features for affect classification is to use deep learning methods to learn latent representations of the input data. These models have been shown to better predict mood, health, and stress, which are all factors in affect, with less error than manually extracted hand-crafted features \cite{Boning}. Deep learning allows for the model to be trained on a large dataset allowing it to find more complex representations that can be better utilized for affect classification \cite{kramer1991nonlinear}. 

Deep learning models commonly employ \textit{fully} supervised learning, which requires human supervision in the form labelled data for both representation learning and classification. However, this reliance on human labelling can reduce the amount of data available to train the model. Additionally, the output labels can make it difficult for the model to generalize across multiple datasets as the differences between the stimuli used in the datasets may elicit different levels of affective response, resulting in vastly different output labels \cite{Affective_reactions}. Combining datasets to create a multi-corpus pipeline is also difficult with supervised learning as the output labels may be carried out with different protocols or standards. Our goal in this work is to reduce the reliance on human supervision by proposing a framework where a significant portion of the processing, i.e. representation learning, can take place in an unsupervised manner, followed by a supervised classification task. This approach should reduce reliance on human supervision and enable the aggregation of several datasets for the representation learning stage of the system, allowing for higher generalizability.

In this paper, an unsupervised solution for representation learning followed by arousal classification using wearable-based biosignals is proposed.Our solution utilizes stacked convolution autoencoders for ECG and EDA representation learning. Autoencoders are unsupervised neural network that attempt to generate output values approximately identical to the inputs successive to an intermediate step where latent representations are learned \cite{Masci_Stacked}. This allows the model to automatically learn important latent representations of the input biosignals without the need for supervised labelling. Next, successive to unsupervised representation learning, we fuse the latent representations and utilize a random forest classifier for classification of low and high arousal. 

The contributions of this work can be summarized as follows: 
\begin{itemize}
    \item We present a novel affective state classification solution consisting of unsupervised multi-modal representation learning using stacked convolutional autoencoders followed by supervised learning of arousal states using a random forest classifier. Our proposed solution benefits from the lack of dependence on user input and supervision throughout the representation learning stage and as a result facilitates the aggregation of datasets, leading to higher generalizability.
    \item We train and test the proposed solution with a aggregated dataset created by combining 3 publicly available datasets, AMIGOS \cite{AMIGOS}, ASCERTAIN \cite{ASCERTAIN}, and MAHNOB-HCI \cite{MAHNOB}, along with CLEAS dataset \cite{Ross_2019} collected in our previous work. Moreover, we compare the performance with a number of baseline techniques for both feature extraction and classification. For feature extraction, we explore a large number of hand-crafted features as well as automatically learned representations using a convolutional neural network (CNN), while for classification we implement a handful of classifiers for comparison. Additionally, we compare the performance of our to a large number of related works in this area.
    \item The results demonstrate the superiority of our proposed approach, outperforming all the baseline techniques as well as the related works, \textit{achieving state-of-the-art} on AMIGOS, ASCERTAIN, CLEAS and MAHNOB-HCI datasets.
    \item Lastly, our analysis shows the added benefit of utilizing a multi-modal approach versus a uni-modal one. Additionally, the added advantage of our unsupervised representation learning method in facilitating an easy aggregation of multiple datasets is demonstrated through comparing our multi-corpus versus single dataset results, in which our multi-corpus approach achieve better performance.
\end{itemize}

The rest of this paper is organized as follows. Section \ref{Related Work} describes the background and related work while section \ref{method} describes our proposed framework utilizing stacked convolutional autoencoders. Section \ref{experiments} gives details on the experiments performed using the proposed solution. Section \ref{results} highlights the results of our framework on the 4 datasets, along with a detailed comparison with other methods, and Section \ref{conclusions} presents a summary and future directions. 

\section{Background and Related work}\label{Related Work}

The term affective computing was first introduced by Rosalind Picard \cite{picard_2000} to describe a new form of human computer interaction where the computer can recognize or influence the emotions of the users. Since its introduction, many studies have looked into how best to recognize emotions, as well as how computers should react to those emotions to enhance the user experience. 

The most widely accepted model for emotion (affect), is the circumplex model proposed by Russel \cite{Russell}. The model utilizes 2 dimensions to describe emotional states, \textit{valence} and \textit{arousal}. Valence refers to how positively or negatively a person is feeling, while arousal refers to how relaxed or stressed they are feeling. This model allows for different emotions to be placed on this circle such that they can be defined as a combination of valence and arousal \cite{Russel&Pratt}.

User-generated information such as facial expressions \cite{sepas2019deep, Sepas-Moghaddam}, gait \cite{etemad2016expert, etemad2014extracting}, speech \cite{fernandez2005classical, tzirakis2018end}, and bio-signals \cite{zhang2019capsule, koelstra2010single}, among others, are then used as inputs to emotion recognition or analysis methods, while the quantified and captured user affective states (e.g. SAM score) are used as outputs. Accordingly, machine learning techniques are often exploited to learn to classify or estimate the target affective states. In the following sub-sections, we present the past works that have used biosignals for affective computing, with a particular focus on works that have utilized the datasets being used in this study. These works can be divided into two categories: uni-modal and multi-modal. While our approach in this paper is multi-modal, specifically using ECG and EDA, to provide a better summary of the work done in this field, we also review other uni-modal techniques.

\subsection{Affective Computing with Uni-modal Biosignals}

Biosignals are the most widely used source of user generated input for machine learning classification of affect. This is due to advancements in the quality and usability of wearable sensors for data collection, along with the correlation between changes in affect, changes in the autonomic nervous system response, and changes in biosignals \cite{Greene}. The biosignals that are typically used for emotion classification are Electroencephalogram (EEG) \cite{koelstra2013fusion}, Electrodermal response (EDA), Electromyography (EMG), and Electrocardiogram (ECG) \cite{Katsigiannis} among others. The use of EDA and ECG based features has been a particularly large focus of study when it comes to classifying the valence and arousal components of emotions. This is due to the fact that these signals have been found to be the most closely correlated with changes in affect \cite{Healey}, and are also relatively easy to record (for example compared to EEG). 

\subsubsection{Classical Machine Learning Techniques}
Gjoreski et al. \cite{Gjoreski2018} used manual feature extraction along with various classical machine learning models to find which models achieve the best performance for classifying arousal from biosignals. They extracted time domain features from ECG and EDA signals from the AMIGOS, ASCERTAIN, and MAHNOB-HCI datasets, separately. With the AMIGOS dataset the best results for ECG features was obtained using a K-Nearest neighbours (KNN) classifier achieving an accuracy of 0.53 while the best result for EDA was with a classifier using AdaBoosting with a decision tree as the base classifier. The best classifier for arousal classification with the ASCERTAIN dataset was a support vector machine (SVM) classifier achieving 0.66 accuracy for both modalities. Lastly, with the MAHNOB-HCI dataset the best performance found for ECG-based features was using an SVM with an accuracy of 0.62 while the best EDA-based classifier was a Naive-Bayes classifier with an accuracy of 0.62. Wiem et al. \cite{wiem2017} also used an SVM classifier, instead utilizing only features extracted from ECG signals in the MAHNOB-HCI dataset. The features were used for binary arousal classification achieving an accuracy of 62\%.

Features extracted from ECG signals were used in \cite{Katsigiannis} for the classification of valence and arousal using SVM classifiers with the DREAMER dataset. Affective data was collected from participants as they were asked to watch videos. Features were extracted by first identifying the PQRST waves within the ECG signal. Frequency based features were also extracted from the Power Spectrum Density (PSD) of the signal. Their work demonstrated the applicability for the use of these features with classical methods for arousal classification achieving an accuracy of 62\% and F1 score of 0.53 for valence and 62\% and 0.58 for arousal.

The capability for EDA to be used to classify both valence and arousal using classical machine learning classifiers was explored in \cite{Greco} where EDA was used to discern 4 levels of arousal and 2 levels of valence. The EDA signals were collected as subjects were stimulated with sounds from the International Affective Digitized Sound System database. Features were then extracted and used with a KNN classifier to obtain 84\% for valence and 77.33\% accuracy for arousal.

\subsubsection{Deep Learning Techniques}
Santamaria et al. \cite{Santamaria} looked at using different feature extraction and arousal classification methodologies with the AMIGOS dataset. They compared manual feature extraction with classical machine learning models, with using a deep CNN for feature representation learning. The input to the CNN was pre-processed ECG and EDA signals. The CNN was comprised of 4 convolutional layers with max-pooling and dropout layers in between them and was compiled using the RMSProp optimizer with a learning rate of 0.0001. The output of the CNN was fed into a Multi-layer Perceptron (MLP) with 4 fully connected layers that produced an arousal classification.  They found that their deep learning approach outperformed the use of classical machine learning methods achieving an accuracy and F1 score of 0.81 and 0.76 when using ECG signals, and 0.71 and 0.67 when using EDA signals. 

Gjoreski et al. \cite{Gjoreski2017} also looked at comparing deep networks with classical machine learning models. In their study they utilized Deep Neural Networks (DNN) with the ASCERTAIN dataset for arousal classification using ECG signals. The DNN contained 7 hidden layers using the ReLU activation with L2 activity regularization. Dropout with a  probability of 0.75 was used between each of the dense layers. The network was trained with the ADAM optimizer with a learning rate of 0.0001. The output from this network utilized the softmax activation function in order to give a class probability distribution for arousal. A random forest, SVM, and decision tree classifier using AdaBoosting classifier were also developed for comparison. The best results were found through using their developed DNN obtaining an accuracy of 0.69. 

Another study that looked at using deep networks for arousal classification was performed by Sicheng et al. \cite{Sicheng}. In the study, a hypergraph learning framework was developed to classify arousal using the ASCERTAIN dataset. The framework produces a correlation between the changes in physiological signals and personality of the subjects, based on the stimuli used in the dataset to evoke emotions. The framework was utilized to classify 2 classes of arousal with ECG and EDA signals, separately. EDA outperformed the ECG classifier obtaining an accuracy of 0.75 compared to 0.72. These results were the best found when using ECG and EDA signals from the ASCERTAIN dataset. 

Instead of classifying valence or arousal, Sarkar et al.  \cite{Sarkar_2019} instead looked at utilizing DNN with only ECG biosignals for expertise and cognitive load classification with the CLEAS dataset. Time and frequency domain features were manually extracted from the ECG signal and then utilized within the DNN to classify binary levels of cogntive load and arousal. The DNN consisted of an input layer accepting the extracted features as input vectors, followed by 7 fully connected dense hidden layers and a final output layer. After each hidden layer, the leaky Rectified Linear Unit (ReLu) activation function was used along with a dropout of 0.5 to prevent over-fitting. The results obtained achieved an accuracy and F1 score of 0.89 and 0.88, and 0.97 and 0.97, respectively. These findings are the current best results for affective state classification using the CLEAS dataset.

In \cite{sarkar2020self, sarkar2020self2}, Sarkar and Etemad proposed a self-supervised method for ECG representation learning in the context of affective computing. The method first used the input signals to generate transformed versions of the data and automatically generated labels for the transformed signals corresponding to the transformation functions. These signals and labels were exploited to train a multi-task CNN, which upon successful training learned generalized representations of the unlabelled ECG signals. Transfer learning was then used to train a supervised CNN for classification of valence and arousal in 4 public datasets including AMIGOS. The arousal classification using the fully supervised model achieved 0.84 accuracy and an F1 score of 0.83 while the self supervised model achieved better results with 0.89 accuracy, a 0.88 F1 score. These are the best results that could be found, to the best of our knowledge, for the AMIGOS dataset. The findings demonstrate the possible benefits when utilizing self supervised learning as opposed to fully supervised methods. 

\subsection{Affective Computing with Multi-modal Biosignals}
In addition to using biosignals as uni-modal inputs for machine learning classification, studies have also used multi-modal inputs as it can utilize the complementary nature that different modalities can present \cite{koelstra2011deap}. 

\subsubsection{Classical Machine Learning Techniques}
The multi-modal use of ECG and EDA signals has been a large area of study. In \cite{das2016emotion} the results of using uni-modal and multi-modal manually extracted features was compared for classification of 3 emotional states; happy, sad, and neutral. The results showed that using the multi-modal features with an SVM classifier outperformed the use of uni-modal features with an average accuracy of 91.62. In \cite{Ross_2019} ECG and EDA time and frequency domain features were extracted from the CLEAS dataset to classify 2 levels of expertise, novice and expert. The study utilized both uni-modal and multi-modal features in a variety of machine learning classifiers. The best result was achieved using multi-modal features with a KNN classifier obtaining an accuracy and F1 score of 0.83 and 0.80. Both studies highlighted the benefits of combining modalities for affective state classification. Anderson et al. \cite{Anderson2017} investigated the multi-modal fusion of EDA and ECG signals with other biosignals. The other signals utilized were EOG, EEG, and photoplethysmogram (PPG). The biosignals were obtained from participants as they participated in a number of stimulating tasks such as listening to music, watching videos, or playing video games. 98 features were extracted from the time and frequency domains of the varying modalities using manual feature extraction. The features were then used within a SVM for arousal classification achieving the best accuracy of 89\% when all modalities were utilized.

A number of datasets have been collected that were subsequently utilized for multi-modal arousal classification. Miranda-Correa et al. \cite{AMIGOS} collected the AMIGOS dataset and then utilized it for arousal classification by manually extracting 77 features from the collected ECG signals, 31 features from the EDA signals. The features were used separately, in addition to being fused to create a multi-modal feature set. The features were used with a Naive-Bayes classifier to classify high and low arousal. F1 scores of 0.55, 0.54, and 0.56 were achieved when using ECG, EDA, and fused features, respectively, demonstrating that fused ECG and EDA signals can achieve better performance than uni-modal features. Subramanian et al. \cite{ASCERTAIN} collected and then utilized  the ASCERTAIN dataset for classification of 2 levels of arousal. They extracted 32 ECG and 31 EDA features for use in machine learning models for valence and arousal classification. SVM and Naive-Bayes classifiers were used with both the uni-modal and multi-modal features. The best results were obtained using a Naive Bayes classifier with fused ECG and EDA features obtaining an F1 score of 0.69. The MAHNOB-HCI dataset was introduced by Soleymani et al. \cite{MAHNOB}. In their preliminary study, 64 ECG features, and 20 EDA features were extracted and subsequently used with an SVM classifier. This method obtained an accuracy of 0.46.

\subsubsection{Deep Learning Techniques}
Deep learning methods have also been used for affective computing with multi-modal biosignals as inputs. Similar to our work, the use of autoencoders for learning representations towards arousal classification have also been investigated. Yang et al. \cite{Yang} utilized an attribute-invariant loss embedded variational autoencoder (VAE) to extract latent representations from ECG, EDA, and EEG signals. These latent representations were used with an SVM classifier to classify binary levels of arousal achieving an accuracy of 0.69 and an F1 score of 0.69. Another method for latent feature extraction for valence and arousal classification was investigated by Liu et al. \cite{liu2016emotion}. In the study a bi-modal Deep Autoencoder was developed to extract latent features from EEG, ECG, and EDA signals from the DEAP dataset. The features were subsequently used by an SVM with a linear kernel to classify valence and arousal, achieving an accuracy of 85\% and 81\% for valence and arousal, respectively.

The benefits of using multi-modal features in deep learning frameworks was further explored by Siddharth et al. \cite{siddharth2019}. In their study, they utilized ECG, EDA, and EEG biosignals from the AMIGOS, DEAP, DREAMER, and MAHNOB-HCI datasets for the classification of valence, arousal, liking, and emotions. Their method utilized a combination of manually extracted statistical features in addition to deep-learning-based features. These feature were obtained by first converting the biosignal time series data to a spectrogram image and then using a pre-trained VGG-16 \cite{VGG} CNN to learn latent features from the image. The features were concatenated used within an Extreme Learning Machine (ELM) for classification. For the AMIGOS, and DEAP datasets, the best performance was obtained using the multi-modal features, showing the potential benefits of using multi-modal data. The DREAMER and MAHBOB-HCI datasets had the best results when using only uni-modal ECG features. The result for the MAHNOB-HCI dataset is noteworthy as it presents the current state-of-the-art performance for the dataset with an accuracy of 0.82 and a F1 score 0f 0.75.

\section{Method} \label{method}
\begin{figure}[!tbp]
\centering
\includegraphics[width=1\linewidth]{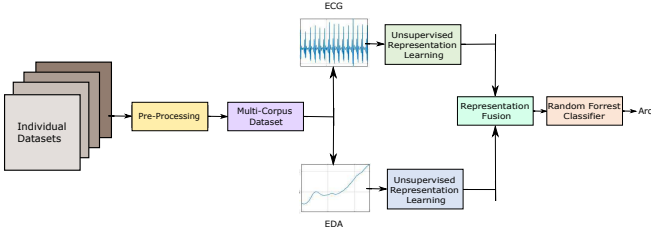}
\caption{Proposed framework for latent biosignal feature representations extraction and subsequent arousal classification.}
\label{fig:Proposed Framework}
\end{figure}

Our goal is to develop a framework for multi-modal affective computing with minimal human supervision. Accordingly, we propose a framework that utilizes stacked convolutional autoencoders for unsupervised ECG and EDA representation learning, and subsequent classification of arousal using a supervised classifier. The method takes filtered and normalized ECG and EDA signals and utilizes 2 separate unsupervised autoencoders, 1 for each modality, to learn latent representations. These representations are then used with a random forest classifier for 2-class classification of arousal. The proposed framework is summarized in Figure \ref{fig:Proposed Framework}. 

\subsection{Pre-Processing}
The Pan-Tompkins algorithm was utilized to filter the raw ECG signals \cite{PanTompkins_1985, Bali_2018}. The Pan-Tompkins algorithm first uses a Butterworth bandpass filter with a passband frequency of 5-15 Hz to reduce the EMG noise, powerline noise, baseline wander and T-wave interference. The filter was applied in both the the forward and reverse directions to achieve zero phase distortion. Following the methods used in \cite{GSR_Guide} the EDA signals were first filtered using a low-pass filter with a cutoff frequency of 1 Hz. High frequency artifacts were then removed with a moving average filter of 100 samples.

\begin{figure}[!tbp]
\centering
\includegraphics[width=0.7\linewidth,trim={10cm 1cm 10cm 0cm, clip}]{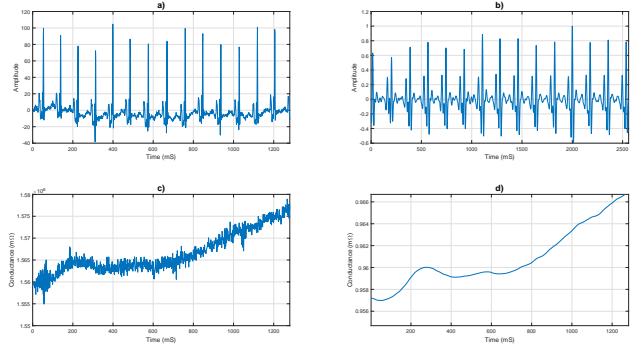}
\caption{Example of \textbf{a)} 10 seconds of raw ECG signal; \textbf{b)} 10 seconds of raw EDA signal; \textbf{c)} 10 seconds of filtered and normalized ECG signal; \textbf{c)} 10 seconds of filtered and normalized EDA signal}
\label{fig:raw and filtered signals}
\end{figure}

After filtering, the ECG and EDA signals were re-sampled to 256 Hz and 128 Hz, respectively. This was done to ensure that the input data from various datasets were the same size. The signals of individual subjects were also normalized to values between 0 and 1 for better use with activation functions within the framework. Examples of the raw ECG and EDA signals along with the filtered and normalized signals are shown in Figure \ref{fig:raw and filtered signals}. 

The ECG and EDA signals were subsequently segmented into 10-second windows to form individual samples. The window size was selected similar to that in \cite{ECG_windowing} for maximum performance while being small enough to better enable real-time applications.

\subsection{Unsupervised Multi-modal Representation Learning}

An autoencoder is an unsupervised learning technique that utilizes backpropagation and takes unlabelled input values \textbf{x}\textsubscript{i}, where \textbf{x}\textsubscript{i} $\in \mathbb{R}$, and attempts to map them to an output, $\hat{\textbf{x}}\textsubscript{i}$, where $\hat{\textbf{x}}\textsubscript{i}$ $\in \mathbb{R}$. The autoencoder is divided into two components, the encoder and the decoder. In the encoder component, the input values \textbf{x}\textsubscript{i} are mapped to a latent representation \textbf{h}\textsubscript{i}. 

The input values are mapped to the latent representation \textbf{h}\textsubscript{i} by using a deterministic function such that \cite{Masci_Stacked}:
\begin{equation}
    \textbf{h}\textsubscript{i} = \sigma\textsubscript{enc}(W\textsubscript{enc}\textbf{x}\textsubscript{i} + b\textsubscript{enc}),
\end{equation}
where $\sigma\textsubscript{enc}$ is the deterministic encoding function, $W\textsubscript{enc}$ is the the weight matrix of the encoding section of the autoencoder and $b\textsubscript{enc}$ is the bias of the encoder \cite{LiuAutoencoder}. The decoder section of the model follows the same approach to map to the output values  $\hat{\textbf{x}}\textsubscript{i}$ given that:
\begin{equation}
     \hat{\textbf{x}}\textsubscript{i} =  \sigma\textsubscript{dec}(W\textsubscript{dec}\textbf{h}\textsubscript{i} + b\textsubscript{dec}),
\end{equation}
where $\sigma\textsubscript{dec}$ is the deterministic decoding function, $W\textsubscript{dec}$ is the the weight matrix of the decoding component, and $b\textsubscript{dec}$ is the bias of the decoder. Through the encoding and decoding functions, the autoencoder seeks to approximate the identity function such that the output values $\hat{\textbf{x}}\textsubscript{i}$ are the same as the input \textbf{x} \textsubscript{i} \cite{SparseAutoencoder}. The autoencoder is trained to minimize the reconstruction error based on the loss function $L(\textbf{x}, \hat{\textbf{x}})$. In our study, the loss function used was the mean squared error such that:
\begin{equation}
    L(\textbf{x}, \hat{\textbf{x}}) = \frac{1}{n} \Sigma_{i=1}^{i=n}(\textbf{x}\textsubscript{i}- \hat{\textbf{x}}\textsubscript{i})^2 .
\end{equation}

The methodology of an autoencoder allows for the latent representations \textbf{h}\textsubscript{i} to be comprised of only features that are the most relevant for the reconstruction of the input values. A sparse autoencoder can be created using the same framework, but introduces a sparsity constraint to the hidden layers \cite{SparseAutoencoder}. By introducing this constraint, the autoencoder is still able to learn latent representations relevant to the structure of the input, even if the number of units in the latent space is large \cite{SparseAutoencoder}. In our proposed method the sparsity constraint is added to the autoencoder through the use of L1 regularization \cite{regularization}. The L1 regularizer is added to the loss function $L$ such that:
\begin{equation}
    L(\textbf{x} , \hat{\textbf{x}} ) = \frac{1}{n} \Sigma_{i=1}^{i=n}(\textbf{x}\textsubscript{i}-\hat{\textbf{x}}\textsubscript{i})^2 + \Sigma_{i=1}^{i=n}(|w\textsubscript{i}|),
\end{equation}
where $w$ is the weight. When multiple hidden layers are combined one after another, the framework is known as a stacked autoencoder. 
In this framework, the input to each subsequent layer is the output from the previous hidden layer, which is the latent representation of the signal generated by that layer \cite{Masci_Stacked}. The framework that is proposed in this study for arousal classification utilizes 2 separate stacked convolutional autoencoders, one for ECG and one for EDA, henceforth denoted by $AE_{ECG}$ and $AE_{EDA}$ respectively. These autoencoders are used to extract latent representations from the output of the \textit{encoder} component, i.e. the central hidden layer. The L1 regularization imposed on this central hidden layer aids with ensuring relevant representations are learned.

\begin{figure}[!tbp]
\centering
\includegraphics[width=0.9\linewidth]{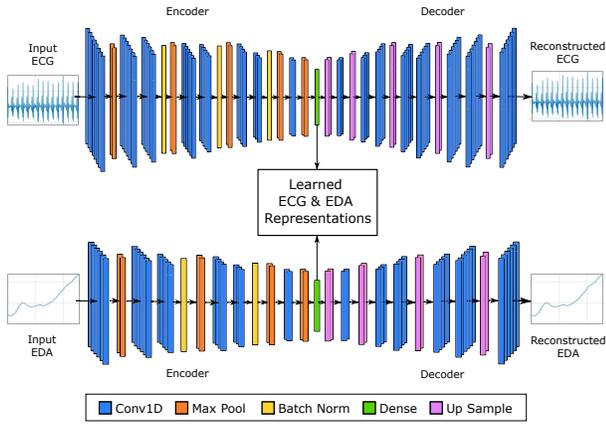}
\caption{The architectures of the proposed autoencoder networks to learn latent representations are presented.}
\label{fig:Autoencoder Framework}
\end{figure}

The proposed autoencoders are depicted in Figure \ref{fig:Autoencoder Framework}. The hyperparameters, namely number of hidden layers, parameters for those layers, and the number of latent representations for both autoencoders are determined systematically through searching a number of possible combinations with the aim of obtaining the best results for arousal classification. 

The encoder component of the $AE_{ECG}$ consists of an input layer followed by 16 hidden layers, and a final fully connected layer. The inputs to the encoder are 10 second ECG segments (2560 samples) as discussed earlier. The first hidden layer performs a 1D convolution on the input of size 128 $\times$ 200 using the ReLU activation function. A max-pooling layer is then utilized to reduce the size of the samples in half. After this layer a number of \textit{convolutional blocks} are used. The \textit{convolutional blocks} consists of 2 1D convolutional layers with different dimensions followed by a batch normalization layer to reduce covariant shift \cite{batchnorm}, and then a max-pooling layer. Within the ECG encoder component this \textit{convolutional block} is repeated 3 times. After the \textit{convolutional blocks} there is a 10 $\times$ 1 1D convolutional layer, a max-pooling layer, and then finally a fully connected dense layer. The dense layer contains 80 units, with an L1 activity regularizer with a value of 10\textsuperscript{-9} that creates a latent feature representation of size 80.

The latent representations learned by the encoder are fed to the decoder component. The decoder consists of 12 hidden layers followed by an output convolutional layer using the ReLU activation function. The hidden layers are comprised of upsampling and convolution layers in the reverse order of the encoder section. The parameters of each layer matches the corresponding encoder layer. The final output from the decoder, and thus the autoencoder as a whole, is a reconstructed ECG segment of the same size as the input (2560). 

The structure of the layers within the $AE_{EDA}$ are similar to that of the developed $AE_{ECG}$. The main differences stem from the fact that the input EDA samples inputted to the encoder section have a shape of 1280 $\times$ 1. The encoder section is comprised of an input layer followed by 12 hidden layer, and a fully connected layer representing the latent EDA features space. The first hidden is a 1D convolution with 32 filters, and a kernel size of 100 $\times$ 1. All the convolutional layers in the $AE_{EDA}$ use the ReLU activation function. This layer is followed by a max-pooling layer to reduce the shape of the sample by half. These layers are followed by 2 convolutional blocks. This is followed by a convolutional layer with 1 filter and a kernel size of 10 $\times$ 1 and a max-pooling layer to reduce the sample to a size of 80 $\times$ 1. Identical to the $AE_{ECG}$ this layer is fed into a fully connected dense layer with 80 units, and an L1 activity regularizer of 10\textsuperscript{-9} giving 80 latent EDA features. 

The decoder section of the $AE_{EDA}$ contains 9 hidden layers and a final convolutional output layer that are arranged to be the inverse of the encoder section. This results in the stacked convolutional autoencoder reconstructing the original EDA sample with it's initial size of 1280 $\times$ 1.

\subsection{Modality Fusion and Classification}\label{arousal classification}

The latent representations learned from the ECG and EDA signals by the respective unsupervised networks are concatenated with a feature-level fusion strategy to form a combined representation for supervised arousal classification with a random forest classifier. Random forest was selected for this application as it is one of the most commonly used supervised classification models for learning latent representations derived from autoencoders for classification tasks in other fields \cite{zhao2017spectral, abraham2018computer, ma2018high, son2018short}.
% These samples were labelled with one of 2 classes, high or low arousal. 

Random forest \cite{Decision_Tree} is an ensemble method consisting of several decision trees that each perform an individual classification which contributes to the final decision. The ensemble decision is made based on a maximum vote among the individual classification results. The number of decision trees used in the random forest, along with the other parameters used, were searched for and determined systematically to achieve the highest accuracy. It was found that a random forest of 100 trees achieved the best results. Bootstrap samples were used when building the decision trees. Class weights were selected to be inversely proportional to the class frequencies within each bootstrap sample.

\section{Experiments}\label{experiments}

\subsection{Datasets}
Four datasets were utilized to enable a multi-corpus evaluation of our proposed solution in order to ensure the approach can work effectively using different wearable devices in varying settings. The four datasets used in this study were AMIGOS \cite{AMIGOS}, ASCERTAIN \cite{ASCERTAIN}, CLEAS \cite{Ross_2019, Sarkar_2019}, and MAHNOB-HCI \cite{MAHNOB}. The following sections provide a detailed description of each dataset, followed by a summary of properties and differences presented in Table \ref{datasets}.  

\begin{table*}[!htbp]
    \centering
    \caption{Summary of the differences between the 4 datasets used.}
    %\begin{tabular}{l|l}
\begin{tabular}{c|cccc}
\toprule
\textbf{Dataset}& \textbf{AMIGOS \cite{AMIGOS}}& \textbf{ASCERTAIN \cite{ASCERTAIN}}& \textbf{CLEAS \cite{Ross_2019}}& \textbf{MAHNOB-HCI \cite{MAHNOB}} \\
\midrule
\textbf{Subjects} & 40 & 58 & 10 & 30\\
\textbf{Stimulus} &  Movie Clips & Movie Clips & Medical Simulation &  Movie Clips, Pictures \\
\textbf{Number of Trials} & 16 & 36 & 2 & 20 \\
\textbf{Duration of Trials} & Varies & 80 seconds & 10 minutes & 12-22 seconds \\
\textbf{ECG System} & 5 channel & 3 channel & 5 channel & 3 channel\\
\textbf{GSR System} & 1 channel & 1 channel & 1 channel & 1 channel\\
\multirow{2}{*}{\textbf{Sampling Rate}} & \multirow{2}{*}{128 Hz} & ECG: 256 Hz & \multirow{2}{*}{500 Hz} & \multirow{2}{*}{256 Hz} \\
& & EDA: 128 Hz & &\\
\textbf{Arousal Rating} & 9-point SAM scale & 7-point scale & 4-point scale & 9-point SAM scale\\
\midrule
%\Xhline{2\arrayrulewidth}
% \bottomrule
    \end{tabular}
    \label{datasets}
\end{table*}

\subsubsection{AMIGOS: A dataset for multi-modal research of affect, personality traits and mood on Individuals and GrOupS \cite{AMIGOS}} \label{AMIGOS Description} 
This dataset contains collected biosignals as well as and Big-Five personality data \cite{Roccas} from 40 participants during short and long emotional videos. The short videos were 250 seconds in length while the long videos were more than 14 minutes long. The participants' EEG, ECG, and EDA signals have been collected with wearable sensors. The ECG system used had 5 channels, while the EDA system had 1 channel. Both signals were captured at a sampling rate of 128 Hz. Arousal values for each video were obtained from the subjects using the 9-point Self-Assessment Manikin (SAM) scale \cite{SAM}.

\subsubsection{ASCERTAIN: A multi-modal databaASe for impliCit pERsonaliTy and Affect recognitIoN \cite{ASCERTAIN}} \label{ASCERTAIN Description} 
This dataset contains the Big-Five personality trait data from 58 participants along with their EEG, ECG, EDA, and facial activity. These data were collected as the participants watched affective movie clips. The average length of the movie clips were 80 seconds. ECG data was captured using a 3-channel system with a sampling rate of 256 Hz, and EDA was single channel with a sampling rate of 128 Hz. Arousal values were elicited from the participants using a 7 point scale from 0 (very boring) to 6 (very exciting). 

\subsubsection{CLEAS: Cognitive Load and Expertise for Adaptive Simulation \cite{Ross_2019, Sarkar_2019}} \label{CLEAS Description} 
We collected this dataset which contains biosignals as well as cognitive load and arousal information acquired during medical simulations. This dataset was collected in collaboration with researchers from Kingston General Hospital \cite{khsc_2020}. Ethics approval was secured from the Queen's University Research Ethics Board. 

A total of 10 participants were recruited for the study, 5 experts (3 male, 2 female) and 5 novices (2 male, 3 female). The participant in the simulation played the role of a trauma team leader put in charge of directing a trauma team on how to best provide care for the patient. The participant was placed in 2 different 10 minute long simulations were developed in which a patient had suffered trauma. A SimMan patient simulator (a mannequin) \cite{SimMan} was used as the patient in these simulations.

Shimmer3 wearable sensors \cite{shimmer} were used to collect ECG, EDA, body temperature and inertial measurement unit (IMU) data during the simulations at . Videos of the simulations were recorded with a Microsoft HoloLens \cite{holo} worn by the participants. %The Hololens was also utilized to place augmented reality medical instrumentation in the room to enhance the visual fidelity of the simulation, in addition to adding visual distractors to manipulate cognitive load. The sensors and instrumentation worn by the participants is shown in \ref{fig:simulation}. The biosignal data were also collected for 2 minutes prior to the simulation in a quiet room in order to collect baseline reference data. The signals were all collected at a sampling rate of 500 Hz. The ECG sensor allows for ECG signals to be measured from four bipolar limb leads and a single chest lead. For the purposes of the study, only the signal obtained from the differential ECG channel between the left arm and right arm leads was used.
The first person videos recorded during the simulations were used during debriefs with the participants successive to completing the simulation. The videos allowed for the participants to review their performance and to annotate their arousal throughout the simulations. The arousal values were collected using a 4 point scale from 1 (calm) to 4 (anxious). 
% The  summarized in Table \ref{datasets}.

\subsubsection{MAHNOB-HCI: A multi-modal Database for Affect Recognition and Implicit Tagging \cite{MAHNOB}} \label{MAHNOB Description}
This dataset contains biosignal data as well as valence and arousal score obtained when showing participants fragments of movies and pictures. The 30 participants in the study were shown the movies and pictures for between 12 and 22 seconds. The valence and arousal data collected were obtained by having the participants annotate their own affective states using the SAM scale. The biosignals obtained during the study were ECG, EEG, EDA, respiration amplitude, and skin temperature. Both the ECG and EDA data were captured with a sampling rate of 256 Hz.

\subsection{Data Labelling}
The learned latent ECG and EDA representations based on the aggregated dataset used in this study are labelled for arousal. Of the 4 datasets, 2 of them, AMIGOS and MAHNOB-HCI, used a 9-point SAM scale for annotating arousal values during the various trials. The ASCERTAIN dataset used a 7-point scale, and the CLEAS dataset used a 4-point scale. To ensure consisted labelling across the multi-corpus the arousal values from each dataset were normalized to values between 1 and 9. This allowed for the datasets to all reflect the 9-point SAM scale. Similar to many works in this area \cite{Gjoreski2017, Gjoreski2018, siddharth2019, Santamaria, Sicheng, wiem2017}, the arousal values were split into two classes: low and high. Following the work done in \cite{koelstra2013fusion}, an arousal value less than 5 was labelled as `low' while an arousal value greater than or equal to 6 was considered high `arousal'.

\subsection{Implementation Details, Training, and Validation}
The baseline machine learning models (discussed later in Section \ref{baseline classifiers}), along with the proposed stacked convolutional autoencoder are developed using scikit-learn \cite{pedregosa2011scikit}, TensorFlow \cite{tensorflow2015} and Keras \cite{chollet2015keras} on a computer with a NVIDIA GeForce RTX 2080 TI graphics card and a Intel Core i7-9700k CPU. The two stacked convolutional autoencoders are trained and validated using a 10-fold validation scheme. Within this validation scheme the aggregated dataset are randomly divided into 10 folds while being stratified such that the percentage of samples in each fold from the individual datasets is representative of the percentage of samples from that dataset in the aggregated dataset. 1 of the 10 folds is left out as a validation set while the other 9 folds are used for training the model. Training is performed using the RMSProp optimizer with the mean squared error (MSE) loss function. A batch size of 32 is used during training with 20 epochs. Early stopping is used by monitoring the validation loss with a patience of 4. The model with the best (lowest) loss value prior to early stopping is saved and subsequently used for multi-modal representation learning. The representations extracted from the training folds are passed to the random forest classifier for training while those from the validation fold are used with the trained random forest classifier for final arousal classification. 

% \section{Evaluation and Comparison} \label{evaluation}
\subsection{Evaluation Metrics} 
To evaluate the efficacy of our proposed framework for arousal classification the results were compared with a variety of other methodologies. The comparisons were based on accuracy and F1 score similar to most other studies in the area \cite{siddharth2019, Santamaria, sarkar2020self, Yang}. To do this the number of True Positive (TP), True Negative (TN), False Positive (FP), and False Negative (FN) classifications were calculated. TP and TN measures were the number of correctly classified arousal classes where, while FP and False Negative FN were defined as the number of incorrect classifications. For the purposes of our study the high arousal class was treated as positive, while the low arousal class was negative. 

\subsection{Manual Feature Extraction and Selection}
We compare the multi-modal representations learned by the unsupervised autoencoders to manually extracted and selected features, generally referred to as hand-crafted features. We select ECG and EDA features that are popular in the field of biosignal analysis, especially those int the field of affective computing \cite{camm1996heart, healey2005detecting, GSR_Guide, Gjoreski2017, Gjoreski2018, MAHNOB, ASCERTAIN}.

In order to extract features from the ECG signals, the Pan-Tompkins algorithm was used to detect the QRS complexes of the ECG waves \cite{PanTompkins_1985,Bali_2018}. This is accomplished by differentiating the filtered ECG signal using a 5-point derivative transfer function in order to gain the QRS slope information. The absolute value of the signal is taken, and then a moving average filter is applied to obtain the wave form features. To detect the R-peaks of the ECG waveform, two threshold values are selected in order to differentiate the peaks from noise. The two threshold values are determined iteratively. If there no peaks are detected within a two second time period then a search-back technique was used to identify any R-peaks that were missed.

12 time domain and 8 frequency domain features are extracted from the pre process ECG signals following the work done in \cite{camm1996heart, healey2005detecting}. The time domain features are extracted from the the distance between two subsequent R-peaks identified in the signal, known as the RR intervals. The frequency domain features are extracted using a power spectrum density (PSD) analysis utilizing a Lomb periodogram \cite{lomb1976least}. These extracted features are summarized in Table~\ref{ECG_Features} in Appendix A.

30 time domain features are extracted from the EDA signals, similar to those extracted in \cite{GSR_Guide}. 
%To extract time domain features, the SCR peaks were first identified..
Time domain features were extracted from 5 components of identified SCR event peaks, the rise time, half recovery time, amplitude, area, and prominence. The SCL of the EDA signal was also found and used to extract time domain features. Next, frequency domain features were extracted successive to performing PSD to obtain the total power estimate among other features. The manually extracted EDA features are summarized in Table~\ref{EDA_Features} in Appendix A. 

Subsequent to feature extraction from the ECG and EDA signals Least Absolute Shrinkage and Selection Operator (LASSO) \cite{tibshirani1996} is used to select the suitable features for arousal classification. LASSO is used to determine the regression coefficients of each feature where the larger the coefficient, the greater it's importance for the classification task. Features that have values close to or equal to zero are not suitable for use in models \cite{tang2014}. The results from using LASSO to select important features is presented and discussed in Section \ref{Learned Representations vs. Hand-Crafted Features}.

\subsection{Baseline Classifiers} \label{baseline classifiers}
Several machine learning models were utilized for comparison with our proposed random forest classifier. These classifiers were evaluated using automatically extracted latent features from the stacked convolutional autoencoders, as well as manually extracted features, separately. The parameters of the machine learning models are tuned empirically successive to iterative experiments to optimize the results for both the use of latent representations of features, and hand crafted features. The details of the 6 models developed are described below. 
\begin{itemize}
    \item \textit{Support Vector Machine (SVM)}: 
    % is a classification technique in which the input features are mapped onto a high dimensional feature space. Hyperplanes are then created to divide the feature space to separate the different classes. This is done such that the margin between the different classes is maximized \cite{Cortes1995}. The mapping of the input features is determined by the kernel selected for the model. 
    Several kernels, namely linear, polynomial with a degree of 2 and 3, and radial basis function (RBF) were evaluated. The best results were obtained with an SVM using an RBF kernel, with a regularization parameter of 0.6, a gamma value (kernel coefficient) equal to 1 divided by the number of features, and class weights inversely proportional to the class frequencies.
    \item \textit{K-Nearest Neighbours (KNN)}: 
    % classifiers determine the class of the input sample based on the class labels of the closest samples in the feature space. The most popular class around the sample is selected as the samples class. The variable k represents the number of neighbours used to determine this classification \cite{KNN}. 
    A \textit{k} value of 7 is found to produce the best results when used in conjunction with weighting the neighbours based on the inverse of the distance to the sample. The Euclidean distance metric is used.
    \item \textit{Decision Tree}: 
    % uses a number of subsequent decision function to split the sample based on its values. The decision functions take input features from a root node, through interior nodes to a final classification \cite{Decision_Tree}. 
    The decision tree (DT) with the best results utilizes entropy for the information gain of evaluating the quality of the split at each node, requires a minimum of 10\% of the samples to split an internal node, uses 90\% of the samples for determining the best split, and has class weights inversely proportional to the class frequencies.
    \item \textit{Multilayer Perceptron (MLP)}: 
    % is a neural network that uses a number of hidden layers containing neurons to take an input and train it to produce an output. 
    The MLP developed for use with the automatically extracted latent features has an input dense layer of 160 neurons. All dense layers utilize the ReLU activation function and are followed by a dropout layer with a coefficient of 0.5. Following the input layer the hidden layers are dense layers with 80, 40, and 40 neurons respectively. The output layer is a dense layer with 1 neuron, and uses the sigmoid activation function. The MLP is optimized with the RMSProp optimizer over 200 epochs minimizing binary cross-entropy loss. For use with the manually extracted features the MLP was modified such that the neurons in each layer are 36, 24, 16, and 1, respectively. 
    %\item \textit{Random Forest Classifier with AdaBoosting (AdaB)}  Of the classifiers developed, the random forest classifier was found to achieve the best performance. To further improve the performance AdaBoosting was used in conjunction with the random forest classifier with with 40 estimators and a learning rate of 0.1 as described in Section \ref{arousal classification}.
    %      \item \textit{Random Forest (RF)}: 
    % % is an ensemble method utilizing a number of decision tree classifiers that each perform a classification from the input features. The class that is classified predominantly by the decision tree classifiers is returned as the classification result of the random forest \cite{Random_forest}. 
    % A forest consisting of 100 trees, with bootstrapping and balanced class weights was found to achieve the optimum results. 
\end{itemize}

\subsection{CNN-based Representation Learning and Classification}
Our proposed framework is also compared with utilizing deep CNNs that learns latent representations of the input signals and utilizes them for arousal classification using fully connected layers. 2 CNNs are developed, an ECG CNN and an EDA CNN. The CNN architectures is similar to the architectures developed for the stacked convolutional autoencoders. For the ECG CNN the first 17 layers are identical to that of the $AE_{ECG}$, while in the EDA CNN the first 13 layers are identical to the $AE_{EDA}$. The convolutional layers of these CNNs reduce the ECG size from 2560 $\times$ 1 to 80 $\times$ 1, and the EDA size from 1280 $\times$ 1 to 80 $\times$ 1. After the convolutional layers, there are a series of 4 fully connected layers containing 80, 40, 20, and 1 neurons respectively. The first dense layer contains L1 activity regularization with a value of $1.0\times10\exp{-9}$. In between every two consecutive dense layers a dropout layer with a rate of 0.5 is utilized. The final dense layer uses a sigmoid activation function while the other dense layers use the ReLU activation function. Both of the CNNs are trained using the Adam optimizer with a learning rate of $1.0\times10\exp{-4}$. The two CNNs (ECG and EDA) are trained for 1500 and 4000 epochs respectively, both with a batch size of 64. Training is stopped early based on the validation loss no longer decreasing after 50 epochs. %The details of the layers and parameters used in the CNNs are shown in Table \ref{CNN details}. 
The outputs from the last dropout layer of each CNN are taken and fused together and fed into a MLP consisting of a dense layer with 40 neurons and using the ReLU activation function, followed by a dropout layer with a ratio of 0.5, and finally a dense layer with 1 unit, using the sigmoid activation function. This MLP was trained using the RMSProp optimizer with a learning rate of $1.0\exp{-3}$, for 200 epochs.

\subsection{Fusion Strategies}
The method through which the ECG and EDA features are fused is also compared for the machine learning models using latent and hand crafted features in addition to the developed CNN. For this comparison only the machine learning models found to have the best performance for automatically learned and hand-crafted features are used. The two methods of fusion compared are feature-level fusion and decision-level fusion. 

For feature-level fusion, the ECG and EDA features are combined prior to arousal classification. Within our proposed framework, this entailed using the respective stacked convolutional autoencoders to extract latent representations from the two modalities. These latent features are then simply concatenated prior to use within a machine learning model for arousal classification. A similar process is followed for the manually extracted features in which they are combined after extraction and selection before use within machine learning models. In the case of the CNN, the representation obtained from the two CNNs after the final convolutional layers are taken and fed into a separate MLP. This MLP is identical, in terms of details and parameters, to the dense layers within the developed CNN. The MLP is trained to take the signal representations and output an arousal class value. 

\begin{figure}[!tbp]
\centering
\includegraphics[width=1\columnwidth]{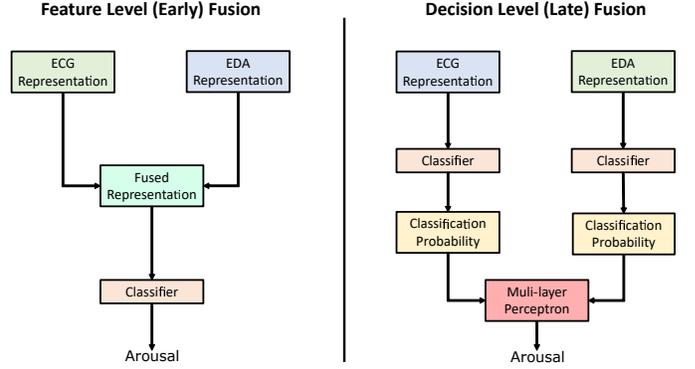}
\caption{Feature level fusion, also known as early fusion, and decision level fusion, known as late fusion, frameworks. }
\label{fig:fusion}
\end{figure}

Decision-level fusion involves classifying arousal based on ECG and EDA features separately, then combining the outputs to form a final classification result. Within the context of using traditional machine learning models with latent and hand-crafted features, this implies that the machine learning models are first trained for each modality separately. The classification probabilities are then taken as the output from the model and fed into an MLP consisting of an input layer of 2 neurons, and an output layer of 1 neuron. Both layers used linear activation function. This method allows for the relative weights of the 2 inputs to be automatically learned to achieve the best results. The frameworks used for feature level and decision level fusion are shown in Figure \ref{fig:fusion}

The best results from comparing feature fusion techniques are taken and compared with the use of uni-modal features. This is accomplished by using the respective ECG and EDA frameworks separately. This is done in order to examine which modality performs the best for arousal classification, and if there is a benefit in combining the two modalities.

\subsection{Single Corpus vs. Multi-Corpus}
One of the intentions behind our unsupervised representation learning framework was easy aggregation of multiple datasets to take advantage of an increased amount of data from an aggregated dataset. However, to evaluate the impact of combining several datasets, we also train the solution with each dataset separately and compare with the multi-corpus approach.

% The proposed unsupervised framework for automated representation learning and arousal classification are designed for use with a multi-corpus dataset. However, the frameworks are also used on each dataset separately for evaluating the impact of our proposed unsupervised representation learning solution in easy aggregation of multiple datasets. 
% This comparison was done to show that the proposed framework has a widespread application and can be used on a variety of datasets that used different experiments and sensors. 
% The results of using the models on separated datasets is compared with the best results from using the combined multi-corpus dataset in order to investigate the benefits that combining datasets for arousal classification may have. 

\begin{table*}%[]
\caption{Classification results using automatically extracted latent features with different machine learning models} 
\centering
  %  \begin{tabular}{l|l}
\begin{tabular}{|>{\centering\arraybackslash}p{1.5cm}|>{\centering\arraybackslash}p{1cm}|>{\centering\arraybackslash}p{1cm}|>{\centering\arraybackslash}p{1cm}|>{\centering\arraybackslash}p{1cm}|>{\centering\arraybackslash}p{1cm}|>{\centering\arraybackslash}p{1cm}|>{\centering\arraybackslash}p{1cm}|>{\centering\arraybackslash}p{1cm}|>{\centering\arraybackslash}p{1cm}|>{\centering\arraybackslash}p{1cm}|}
\toprule        
% \Xhline{2\arrayrulewidth}
         \multirow{2}{*}{\textbf{Classifier}} & \multicolumn{2}{|c|}{\textbf{\makecell{AMIGOS}}} &  \multicolumn{2}{|c|}{\textbf{\makecell{ASCERTAIN}}} & \multicolumn{2}{|c|}{\textbf{\makecell{CLEAS}}}&
         \multicolumn{2}{|c|}{\textbf{\makecell{MAHNOB-HCI}}}&
          \multicolumn{2}{|c|}{\textbf{\makecell{Overall}}}\\
         \cline{2-11}
         & Acc. & F1 & Acc. & F1 & Acc. & F1 & Acc. & F1 & Acc. & F1\\
\toprule
SVM & 0.66 & 0.78 & 0.67 & 0.80 & 0.69 & 0.56 & 0.45 & 0.56 & 0.62 & 0.75\\
\hline
KNN & 0.85 & 0.89 & 0.72 & 0.80 & 1.0 & 1.0 & 0.73 & 0.67 & 0.81 & 0.84 \\
\hline
DT  & 0.81 & 0.86 & 0.62 & 0.72 & 0.88 & 0.78 & 0.53 & 0.57 & 0.74 & 0.79\\
\hline
MLP  & 0.69 & 0.80 & 0.67 & 0.80 & 0.69 & 0.57 & 0.45 & 0.56 & 0.64 & 0.76\\
\hline
\textbf{RF} & \textbf{0.93} & \textbf{0.95} & \textbf{0.79} & \textbf{0.86} & \textbf{0.99} & \textbf{0.98} & \textbf{0.83} & \textbf{0.76} & \textbf{0.89} & \textbf{0.91}\\ 
\bottomrule    
\end{tabular}
\label{Classifier Model Latent Feature Results}
\end{table*}

\subsection{Other Works and State-of-the-Art}
We compare our results to related works in arousal classification for each of the datasets used in this study, described in Section \ref{Related Work}. For a fair comparison, only studies that utilize ECG and/or EDA signals are used. Additionally, we compare to current state-of-the-art results for each individual dataset. Specifically, we compare to \cite{sarkar2020self} for the AMIGOS datasets, \cite{Sicheng} for the ASCERTAIN dataset, and \cite{siddharth2019} for the MAHNOB-HCI dataset. For the CLEAS dataset, where there are not any related works on arousal classification, our results are compared with related works that have performed other affective state classification tasks. 

\section{Results and Discussions} \label{results}
% \subsection{Arousal Classification using Latent Features from Stacked Autoencoders}

\subsection{Performance}\label{performance} 

In this section we explore the performance of the unsupervised autoencoder methodology for developing latent representations in addition to the performance of the supervised arousal classification models.

To evaluate the impact of the supervised classifier in our proposed framework, we utilize the multi-modal representations learned from the proposed unsupervised autoencoders for classification with 5 supervised methods (as discussed in Section \ref{baseline classifiers}). The results are presented in Table \ref{Classifier Model Latent Feature Results}. It can be seen that the random forest classifier obtained the best results on every dataset with an accuracy/F1 score of 0.93/0.95, 0.79/0.86, 0.99/0.98, and 0.83/0.76 for AMIGOS, ASCERTAIN, CLEAS, and MAHNOB-HCI, respectively, and consequently the best overall results with an accuracy of 0.89 and an F1 score of 0.91.

When comparing the performance of the machine learning models used, the performances are ranked in the order of SVM, MLP, KNN, decision tree, and random forest. It is particularly interesting to note the difference in performance between the decision tree, and the random forest as it demonstrates the advantages of using ensemble learning methods. The use of a decision tree classifier achieves an accuracy and F1 score of 0.74 and 0.79, while the random forest classifier with 100 decision trees greatly increases the performance to an accuracy of 0.89 and an F1 score of 0.91. These findings support the use of ensemble learning using the learned multi-modal latent representations for arousal classification.

\subsection{Representation Learning vs. Feature Extraction Techniques} \label{Learned Representations vs. Hand-Crafted Features}

To investigate the impact of the unsupervised representation learning achieved using our proposed framework, we compare the discriminability of the representations to hand-crafted features manually extracted and selected from the ECG and EDA signals. 
% \subsubsection{Feature Selection}\hfill\\
Using LASSO on the extracted features, 11 ECG features and 25 EDA features are found to have non-zero regression coefficient. These features are considered to be important and thus are used within the developed machine learning models for classification.

The EDA feature with the greatest regression coefficient, and thus the greatest importance for arousal classification, is found to be the standard deviation of the half recovery time with a regression coefficient of 0.3083. The features with the next highest importance are those from the rise time of SCR events, followed by the amplitude of SCR events and then the area of the SCR events. The lowest regression coefficients within the EDA important features are those taken from the skin conductance level. This indicates that the SCR events within an EDA signal are likely applicable for the classification of arousal while the SCL of the signal may be less useful. None of the frequency domain features are found to be of importance for this classification task. 

Although more EDA features are found to be important for the classification of arousal, the regression coefficients of the important ECG features were higher than that of EDA. This finding suggests that the ECG features may be more discriminative for detection of arousal than EDA. The highest regression coefficient is 0.8861 for the minimum RR interval. 3 other features based on RR intervals are found to be important: RR interval maximum, difference, and standard deviation with regression coefficients of 0.2635, 0.2341, and 0.2211, respectively. This suggests that the RR peaks are the most important feature that can be extracted from the ECG signal for the purpose of this arousal classification. %, which is in line with other studies in this area \cite{}.

% \subsubsection{Performance}\hfill\\

The features that are selected using LASSO are used as inputs to the same machine learning models developed for use with our proposed unsupervised model. As was the case with the learned latent representations, the best overall results when using hand-crafted features was obtained using the random forest, achieving an accuracy of 0.66 and an F1 score of 0.72. The machine learning models using automatically extracted latent features were also compared with using a deep convolutional neural network for arousal classification. 2 separate CNN were trained using ECG and EDA signals separately.

% These findings show that, for this classification task, the latent features found using stacked convolutional autoencoder are better input features for machine learning models than manually extracted features, in addition to requiring less pre-processing. 

%As noted previously, the use of latent features within an AdaBoosted random forest classifier outperformed the use of more traditional hand-crafted features. Comparing the ROC curve and confusion matrix results of using latent feature extraction prior to arousal classification, with the results of using CNN for arousal classification, it can be seen that the autoencoder framework achieves a higher AUC and better classification accuracies for both arousal classes a seen in the confusion matrices. 

\begin{figure}[!tbp]
\centering
\includegraphics[width=\linewidth,trim={0.5cm 0cm 0cm 0cm,clip}]{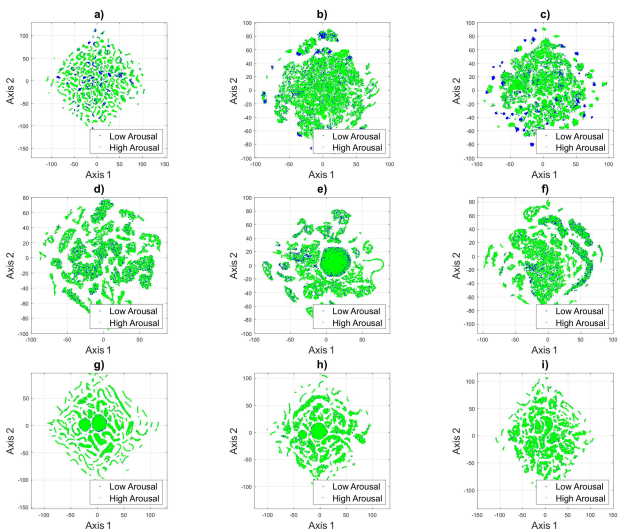}
\caption{tSNE of \textbf{a)} ECG representation learned from the ECG autoencoder; \textbf{b)} EDA representation learned from the EDA autoencoder; \textbf{c)} combined ECG and EDA representations from autoencoders; \textbf{d)} manually extracted ECG features; \textbf{e)} manually extracted EDA features; \textbf{f)} combined manually extracted ECG and EDA features; \textbf{d)} ECG representation learned from CNN; \textbf{b)} EDA representation learned from CNN; \textbf{c)} combined ECG and EDA representations learned from CNN}
\label{tSNE}
\end{figure}

\begin{table*}[!tbp]
\caption{Classification results using uni-modal and multi-modal datasets fused with different methodologies for arousal classification using latent and hand-crafted features with machine learning models, and a deep CNN.} 
    \centering
     
  %  \begin{tabular}{l|l}

\resizebox{\textwidth}{!}{\begin{tabular}{|c|c|>{\centering\arraybackslash}p{0.75cm}|>{\centering\arraybackslash}p{0.75cm}|>{\centering\arraybackslash}p{0.75cm}|>{\centering\arraybackslash}p{0.75cm}|>{\centering\arraybackslash}p{0.75cm}|>{\centering\arraybackslash}p{0.75cm}|>{\centering\arraybackslash}p{0.75cm}|>{\centering\arraybackslash}p{0.75cm}|>{\centering\arraybackslash}p{0.75cm}|>{\centering\arraybackslash}p{0.75cm}|}
\cline{3-12}  
          \multicolumn{2}{c|}{}& 
          \multicolumn{2}{|c|}{\textbf{AMIGOS}}& \multicolumn{2}{|c|}{\textbf{ASCERTAIN}} & \multicolumn{2}{|c|}{\textbf{CLEAS}}& \multicolumn{2}{|c|}{\textbf{MAHNOB}}& \multicolumn{2}{|c|}{\textbf{Overall}}\\
         \midrule
         Method & \makecell{Modality/ \\Fusion} & Acc. & F1 & Acc. & F1 & Acc. & F1& Acc. & F1 & Acc. & F1\\
\toprule
\multirow{4}{*}{\textbf{\makecell{Hand-crafted \\ Features}}} & ECG & 0.63 & 0.73 & 0.57 & 0.67 & 0.65 & 0.42 & 0.54 & 0.39 & 0.59 & 0.65\\
\cline{2-12}
& EDA & 0.62 & 0.75 & 0.66 & 0.79 & 0.88 & 0.67 & 0.61 & 0.43 & 0.64 & 0.71\\
\cline{2-12}
& \makecell{Decision-\\Level} & 0.66 & 0.79 & 0.68 & 0.81 & 0.90 & 0.77 & 0.59 & 0.45 & 0.65 & 0.73\\
\cline{2-12}
& \makecell{Feature-\\Level} & 0.66 & 0.76 & 0.67 & 0.79 & 0.91 & 0.77 & 0.64 & 0.42 & 0.66 & 0.72\\
\hline
\multirow{4}{*}{\textbf{\makecell{Learned CNN\\Representations}}} & ECG & 0.86 & 0.89 & 0.79 & 0.85 & 0.84 & 0.68 & 0.73 & 0.66 & 0.82 & 0.85\\
\cline{2-12}
& EDA & 0.67 & 0.77 & 0.65 & 0.77 & 0.60 & 0.36 & 0.58 & 0.58 & 0.65 & 0.74\\
\cline{2-12}
& \makecell{Decision-\\Level} & 0.88 & 0.91 & 0.79 & 0.86 & 0.84 & 0.67 & 0.75 & 0.67 & 0.85 & 0.87\\
\cline{2-12}
& \makecell{Feature-\\Level} & 0.88 & 0.91 & 0.79 & 0.85 & 0.82 & 0.65 & 0.75 & 0.68 & 0.84 & 0.87\\
\hline
\multirow{4}{*}{\textbf{\makecell{Ours\\(Autoencoders)}}} & ECG & 0.85 & 0.89 & 0.65 & 0.73 & 0.88 & 0.77 & 0.67 & 0.62 & 0.78 & 0.82\\
\cline{2-12}
& EDA & 0.64 & 0.72 & 0.59 & 0.69 & 0.90 & 0.79 & 0.58 & 0.48 & 0.63 & 0.68\\
\cline{2-12}
& \makecell{Decision-\\Level} & 0.76 & 0.80 & 0.60 & 0.67 & 0.92 & 0.80 & 0.68 & 0.57 & 0.73 & 0.75\\
\cline{2-12}
& \makecell{Feature-\\Level} & \textbf{0.93} & \textbf{0.95} & \textbf{0.79} & \textbf{0.86} & \textbf{0.99} & \textbf{0.98} & \textbf{0.83} & \textbf{0.76} & \textbf{0.89} & \textbf{0.91}\\
 \bottomrule
\end{tabular}}
\label{Modality Results}
\end{table*}

\subsection{Feature Space Exploration}
We perform \textit{t}-SNE \cite{maaten2008} on the uni-modal and multi-modal latent representations learned by our proposed unsupervised solution, for visualization. For comparison we also perform \textit{t}-SNE on uni-modal and multi-modal manually extracted features and latent representations learned by CNN.

\textit{t}-SNE was performed using 10000 iterations with a perplexity of 30, and learning rate of 10. The outcome is shown in Figure \ref{tSNE}. The figure illustrates that the learned representations for ECG, EDA, and multi-modal are visibly more separable compared to hand-crafted features and learned representations from CNN, further validating the advantage of using our proposed solution. Additionally, it can be observed that the latent representations learned from ECG exhibit an enhanced separability compared to that of EDA, indicating that ECG is likely to be a more discriminative modality for arousal detection. This could lead to the ECG features being a better modality for arousal classification than EDA.

\subsection{Impact of Multi-modality and Fusion}

We evaluate the impact using both ECG and EDA modalities (multi-modal) versus the two modalities individually (uni-modal). Moreover, we evaluate the performance when a feature-level fusion strategy is used for the learned representations of the two modalities in our solution compared to decision-level fusion. Lastly, we perform the same comparisons for hand-crafted features and CNN-based learned representations. 
% Decision-level fusion was also used to determine the best results for comparison from the methods using manually extracted features with an AdaBoosted random forest classifier and the architecture using latent CNN features. In order to determine the weights for the 2 modalities for decision-level fusion a simple MLP was developed with the probabilities from classifiers using the separate modalities as the input, and arousal as the output. 
% The results obtained using this MLP for decision-level fusion is compared to the results using feature-level fusion in Table \ref{Modality Results}. The results for latent features from the stacked convolutional autoencoder and manually extracted features show that feature-level fusion has better performance than decision-level fusion. When using the CNN architecture for extracting latent features for subsequent use in an MLP the results of decision-level fusion perform slightly better than feature-level fusion.
The results of this analysis is are presented in Table \ref{Modality Results}. 
Looking at the uni-modal results, it is observed that EDA features prove more discriminative when hand-crafted features are used, when automatically learned representations are employed (CNN-based and ours) ECG is a more suitable modality for arousal classification compared to EDA. 
% This finding correlates with those found by using \textit{t}-SNE to explore the feature space as ECG was found to be more discriminitive in terms of arousal classification.
Moreover, the results show that, as expected, multi-modal approaches (both feature-level and decision-level fusion) outperform the use of uni-modal features for the hand-crafted features, CNN-based representations, and the unsupervised learned representations (our method). 
Lastly, the results illustrate that while for hand-crafted features and CNN-based representations, feature-level and decision-level fusion strategies perform on-par with one another, for our unsupervised learned representations, feature-level fusion performs noticeably better than decision-level fusion.

\begin{table*}[!tbp]
\caption{Classification results using multi-corpus and separated datasets for arousal classification using latent and hand-crafted features with machine learning models, and a deep CNN.} 
    \centering

\resizebox{\textwidth}{!}{\begin{tabular}{|c|c|>{\centering\arraybackslash}p{0.75cm}|>{\centering\arraybackslash}p{0.75cm}|>{\centering\arraybackslash}p{0.75cm}|>{\centering\arraybackslash}p{0.75cm}|>{\centering\arraybackslash}p{0.75cm}|>{\centering\arraybackslash}p{0.75cm}|>{\centering\arraybackslash}p{0.75cm}|>{\centering\arraybackslash}p{0.75cm}|>{\centering\arraybackslash}p{0.75cm}|>{\centering\arraybackslash}p{0.75cm}|}
\cline{3-12}  
          \multicolumn{2}{c|}{}& 
          \multicolumn{2}{|c|}{\textbf{AMIGOS}}& \multicolumn{2}{|c|}{\textbf{ASCERTAIN}} & \multicolumn{2}{|c|}{\textbf{CLEAS}}& \multicolumn{2}{|c|}{\textbf{MAHNOB}}& \multicolumn{2}{|c|}{\textbf{Overall}}\\
         \midrule
         Method & Dataset(s) & Acc. & F1 & Acc. & F1 & Acc. & F1& Acc. & F1 & Acc. & F1\\
\toprule
\multirow{2}{*}{{\textbf{\makecell{Hand-crafted\\Features}}}} & Separated & 0.67 & 0.77 & 0.66 & 0.79 & 0.91 & 0.77 & 0.64 & 0.41 & 0.67 & 0.72\\
\cline{2-12}
& Multi-Corpus & 0.66 & 0.76 & 0.67 & 0.79 & 0.91 & 0.77 & 0.64 & 0.42 & 0.66 & 0.72\\
\hline
\multirow{2}{*}{{\textbf{\makecell{Learned CNN\\Representations}}}} & Separated & 0.67 & 0.75 & 0.68 & 0.81 & 0.86 & 0.55 & 0.59 & 0.38 & 0.66 & 0.73\\
\cline{2-12}
& Multi-Corpus & 0.88 & 0.91 & 0.79 & 0.85 & 0.84 & 0.67 & 0.75 & 0.67 & 0.85 & 0.87\\
\hline
\multirow{2}{*}{{\textbf{\makecell{Ours\\(Autoencoders)}}}} & Separated & 0.89 & 0.92 & 0.76 & 0.82 & 0.99 & 0.98 & 0.78 & 0.74 & 0.85 & 0.88\\
%0.93 & 0.95 & 0.78 & 0.85 & 0.99 & 0.98
\cline{2-12}
& Multi-Corpus & \textbf{0.93} & \textbf{0.95} & \textbf{0.79} & \textbf{0.86} & \textbf{0.99} & \textbf{0.98} & \textbf{0.83} & \textbf{0.76} & \textbf{0.89} & \textbf{0.91}\\
 \bottomrule
\end{tabular}}
\label{Corpus Results}
\end{table*}

\begin{table*}[!tbp]
    \centering
    \caption{Comparison of Classifier Results using AMIGOS Dataset}
  %  \begin{tabular}{l|l}
\begin{tabular}{|c|c|c|c|c|c|c|}
\toprule        
% \Xhline{2\arrayrulewidth}
         \textbf{Ref.} & \textbf{Year} & \textbf{Biosignals} & \textbf{\makecell{Features/ \\ Representations}}   & \textbf{Classifier} &   \textbf{Acc.} &  \textbf{F1} \\
        % \Xhline{2\arrayrulewidth}
\toprule
\multirow{2}{*}{Miranda-Correa et al. \cite{AMIGOS}} & \multirow{2}{*}{2018} & ECG & \multirow{2}{*}{Hand-crafted} & \multirow{2}{*}{GNB} & - & 0.55 \\
\cline{3-3} \cline{6-7}
&& EDA & & & - & 0.54 \\
\cline{3-3} \cline{6-7}
&& ECG, EDA, EEG &  &  & - & 0.56 \\
\hline
\multirow{2}{*}{Gjoreski et al. \cite{Gjoreski2018}} & \multirow{2}{*}{2018} & ECG & \multirow{2}{*}{Hand-crafted}&  KNN & 0.53 & - \\
\cline{3-3} \cline{5-7}
&& EDA & & AdaBoost DT & 0.56 & - \\
\hline
Yang et al. \cite{Yang} & 2019 & ECG, EDA, EEG & VAE & SVM & 0.69 & 0.69 \\
\hline
\multirow{3}{*}{Siddharth et al. \cite{siddharth2019}} & \multirow{3}{*}{2019} & ECG &\multirow{3}{*}{\makecell{CNN}} & \multirow{3}{*}{ELM} &  0.83 & 0.76 \\
\cline{3-3} \cline{6-7}
&& EDA &  & & 0.81 & 0.74 \\
\cline{3-3} \cline{6-7}
&& ECG, EDA, EEG & & & 0.83 & 0.76 \\
\hline
\multirow{2}{*}{Santamaria et al. \cite{Santamaria}}& \multirow{2}{*}{2019} & ECG & \multirow{2}{*}{CNN} & \multirow{2}{*}{MLP} &  0.81 & 0.76 \\
\cline{3-3} \cline{6-7}
 && EDA &  & & 0.71 & 0.67 \\
 \hline
\multirow{2}{*}{Sarkar \cite{sarkar2020self}} & \multirow{2}{*}{2020} & \multirow{2}{*}{ECG} & \makecell{Fully Supervised\\CNN} & \multirow{2}{*}{MLP} & 0.84 & 0.83\\
\cline{4-4} \cline{6-7}
&&& \makecell{Self-Supervised\\CNN} & & 0.89 & 0.88 \\
 \hline
\textbf{Ours} & 2020 & \textbf{ECG, EDA} & \textbf{\makecell{Convolutional \\ Autoencoder}} & \textbf{RF} & \textbf{0.93} & \textbf{0.95} \\
   %      \Xhline{2\arrayrulewidth}
\bottomrule    
\end{tabular}
    \label{AMIGOS Comparison}
\end{table*}

\begin{table*}[!tbp]
    \centering
     
    \caption{Comparison of Classifier Results using ASCERTAIN Dataset}
  %  \begin{tabular}{l|l}

\begin{tabular}{|c|c|c|c|c|c|c|}
\toprule        
% \Xhline{2\arrayrulewidth}
         \textbf{Ref.} &   \textbf{Year} &  \textbf{Biosignals} & \textbf{\makecell{Features/ \\ Representations}}  & \textbf{Classifier} &  \textbf{Acc.} &  \textbf{F1} \\
        % \Xhline{2\arrayrulewidth}
\toprule
\multirow{3}{*}{Subramanian et al. \cite{ASCERTAIN}} & \multirow{3}{*}{2016} & ECG &\multirow{3}{*}{Hand-crafted} & \multirow{3}{*}{NB}& - & 0.59\\
\cline{3-3} \cline{6-7}
&& EDA &  & & - & 0.66 \\
\cline{3-3} \cline{6-7}
&& ECG, EDA &  & & - & 0.69 \\
\hline
Gjoreski et al.\cite{Gjoreski2017} & 2017 & ECG & Hand-crafted & DNN &  0.69 & - \\
\hline
\multirow{2}{*}{Gjoreski et al. \cite{Gjoreski2018}} & \multirow{2}{*}{2018} & ECG & \multirow{2}{*}{Hand-crafted} & \multirow{2}{*}{SVM}  & 0.66 & - \\
\cline{3-3} \cline{6-7}
&& EDA & & & 0.66 & - \\
\hline
\multirow{2}{*}{Sicheng et al. \cite{Sicheng}} & \multirow{2}{*}{2018} & ECG & \multirow{2}{*}{Hand-crafted} & \multirow{2}{*}{Hypergraph Learning} &  0.72 & - \\
\cline{3-3} \cline{6-7}
&& EDA & & &  0.75 & - \\
\hline
\textbf{Ours} & 2020 & \textbf{ECG, EDA} & \textbf{\makecell{Convolutional \\ Autoencoder}} &  \textbf{RF} & \textbf{0.79} & \textbf{0.86} \\
   %      \Xhline{2\arrayrulewidth}
\bottomrule    
\end{tabular}
    \label{ASCERTAIN Comparison}
\end{table*}

\subsection{Individual vs. Multi-corpus}
As discussed earlier, in our experiments, 4 datasets are combined to create a single aggregated dataset. In order to investigate the impact of combining the 4 datasets, we compare the outcome of the 3 methodologies (hand-crafted features, CNN-based representations, and ours) when individual datasets are used to train and test the models as opposed to the aggregated dataset. The results presented in Table \ref{Corpus Results} illustrate that while hand-crafted features show very close performance for single datasets and the multi-corpus approach, the CNN-based method and our framework both show a clear advantage to using a larger aggregated dataset. This observation is expected as the quality or type of hand-crafted features do not change with more data, i.e. they remain identical from one dataset to the other. Meanwhile, with different training data, learned representations change, and with a multi-corpus approach for training, their generalization and discriminability improves. Although the use of multi-corpus increased the performance of our method, it is important to note that the use of separate datasets still perform well compared to other methods.

\begin{table*}[!htbp]
    \centering
     
    \caption{Comparison of Classifier Results using CLEAS Dataset}
  %  \begin{tabular}{l|l}
\begin{tabular}{|c|c|c|c|c|c|c|c|}
\toprule        
% \Xhline{2\arrayrulewidth}
         \textbf{Ref.} & \textbf{Year} & \textbf{Classification Task} & \textbf{Biosignals} & \textbf{\makecell{Features/ \\ Representations}} & \textbf{Classifier} & \textbf{Acc.} &  \textbf{F1} \\
        % \Xhline{2\arrayrulewidth}
\toprule
Ross et al. \cite{Ross_2019} & 2019 & Expertise & ECG, EDA & Hand-crafted & KNN & 0.83 & 0.80\\
\hline
\multirow{2}{*}{Sarkar et al. \cite{Sarkar_2019}} & \multirow{2}{*}{2019} & Cognitive Load & \multirow{2}{*}{ECG} & \multirow{2}{*}{Hand-crafted} & \multirow{2}{*}{DNN} & 0.89 & 0.88 \\
\cline{3-3} \cline{7-8}
&& Expertise & & & & 0.97 & 0.97\\
\hline
\textbf{Ours} & 2020 & Arousal & \textbf{ECG, EDA} & \textbf{\makecell{Convolutional \\ Autoencoder}}& \textbf{RF} & \textbf{0.99} & \textbf{0.98} \\
   %      \Xhline{2\arrayrulewidth}
\bottomrule    
\end{tabular}
    \label{CLEAS Comparison}
\end{table*}

\begin{table*}[!htbp]
    \centering
    \caption{Comparison of Classifier Results using MAHNOB-HCI Dataset}
  %  \begin{tabular}{l|l}
\begin{tabular}{|c|c|c|c|c|c|c|}
\toprule        
% \Xhline{2\arrayrulewidth}
         \textbf{Ref.} & \textbf{Year} & \textbf{Biosignals} & \textbf{\makecell{Features/ \\ Representations}}   & \textbf{Classifier} & \textbf{Acc.} &  \textbf{F1} \\
        % \Xhline{2\arrayrulewidth}
\toprule
Soleymani et al. \cite{MAHNOB} & 2010 & ECG, EDA & Hand-crafted & SVM &  0.46 & - \\
\hline
Wiem et al. \cite{wiem2017} & 2017 & ECG & Hand-crafted & SVM &  0.62 & - \\
\hline
\multirow{2}{*}{Gjoreski et al. \cite{Gjoreski2018}} & \multirow{2}{*}{2018} & ECG & \multirow{2}{*}{Hand-crafted} & SVM & 0.62 & - \\
\cline{3-3} \cline{5-7}
&& EDA & & NB &  0.62 & - \\
\hline
\multirow{3}{*}{Siddharth et al. \cite{siddharth2019}} & \multirow{3}{*}{2019} & ECG & \multirow{3}{*}{\makecell{CNN}} & \multirow{3}{*}{ELM} &  0.79 & 0.74 \\
\cline{3-3} \cline{6-7}
&& EDA &  & & 0.82 & 0.75 \\
\cline{3-3} \cline{6-7}
&& ECG, EDA, EEG & & & 0.81 & 0.71 \\
\hline
\textbf{Ours} & \textbf{2020} & \textbf{ECG, EDA} & \textbf{\makecell{Convolutional \\ Autoencoder}} & \textbf{RF} & \textbf{0.83} & \textbf{0.76} \\
   %      \Xhline{2\arrayrulewidth}
\bottomrule    
\end{tabular}
    \label{MAHNOB Comparison}
\end{table*}

% \begin{table}[t]
%     \centering
%     \caption{The average and standard deviations of the accuracies and F1 scores for signal transformation recognition across the four datasets are presented.}
%     \begin{tabular}{l|c|c}
%     \hline
%     \textbf{\multirow{2}{*}{Transformation}} & \multicolumn{2}{c}{\textbf{All datasets combined}} \\ \cline{2-3}
%         & Acc. & F1  \\ \hline \hline
%         \multirow{1}{*}{\textbf{Original}}  & $0.980\pm0.003$ & $0.927\pm0.007 $\\
%         \multirow{1}{*}{\textbf{Noise Addition}} & $0.995\pm0.000$ & $0.979\pm0.003 $ \\
%         \multirow{1}{*}{\textbf{Scaling}} & $0.982\pm0.003$ & $0.932\pm0.010 $ \\
%         \multirow{1}{*}{\textbf{Temporal Inversion}} & $0.998\pm0.000$ & $0.992\pm0.004 $ \\
%         \multirow{1}{*}{\textbf{Negation}} & $0.998\pm0.000$  & $0.990\pm0.000$ \\
%         \multirow{1}{*}{\textbf{Permutation}} & $0.998\pm0.000$  & $0.989\pm0.003$ \\
%         \multirow{1}{*}{\textbf{Time-warping}} & $0.997\pm0.003$  & $0.992\pm0.006$ \\ \Xhline{1pt}
%         \multirow{1}{*}{\textbf{Average}} & $0.992\pm0.001$  & $0.972\pm0.005$ \\ \hline
%     \end{tabular}
%     \label{tab:ssl_result_signal_combined}
% \end{table}

\subsection{Comparison to Other Works and State-of-the-Art}
We compare our results to other related works that utilized the 4 datasets used in this study. Table \ref{AMIGOS Comparison} presents the results for the AMIGOS dataset where our proposed framework outperforms the state-of-the-art. The best result on this dataset was previously achieved in \cite{sarkar2020self}, where a self-supervised approach was used for ECG-based arousal detection with an accuracy and F1 of 0.89 and 0.88 respectively. Our solution outperformed this work with an accuracy and F1 score of 0.93 and 0.95, achieving respective gains of 0.04 and 0.07 for accuracy and F1.

% It is most notable to compare our results with the best previous results using feature extraction, autoencoder, and CNN techniques. The state-of-the-art performance using manual feature extraction was found in \cite{Gjoreski2017} with an accuracy of 0.56. That result was worse than the results found in our study with any methodology. When comparing to other results using autoencoders, as studied in \cite{Yang}, our method greatly outperforms theirs which had an accuracy and F1 score of 0.69 and 0.69, while using one less modality. The best result using deep learning utilized was in \cite{sarkar2020self}, which utilized a CNN and had similar accuracy and F1 to our results using CNN. However, our autoencoder approach still had the best performance with an accuracy and F1 score of 0.93 and 0.95.

%The comparison between our results, and the related works utilizing the ASCERTAIN dataset are shown in Table \ref{ASCERTAIN Comparison}. 
%The best state-of-the-art result when using manual feature extraction was found in \cite{Gjoreski2018} with an accuracy of 0.66. Our results using manual feature extraction slightly outperform this result with an accuracy of 0.67 and an F1 score of 0.79. To the best of our knowledge there has not been a study with the ASCERTAIN dataset using an autoencoder approach. 
The comparison between our results, and the related works utilizing the ASCERTAIN dataset are shown in Table \ref{ASCERTAIN Comparison}. The state-of-the-art result was found in \cite{Sicheng} with an accuracy of 0.75 using EDA alone with manual feature extraction and subsequent classification with hypergraph learning. Our proposed unsupervised learning framework achieves better results on ASCERTAIN with an accuracy of 0.79 and F1 score of 0.86, obtaining a gain in accuracy of 0.04.

Although the CLEAS dataset has not been previously used for arousal classification, we compare our results to related works that have utilized the dataset for other classification tasks as shown in Table \ref{CLEAS Comparison}. The state-of-the-art results for classifying cognitive load and expertise were found in \cite{Sarkar_2019} with an accuracy and F1 score of 0.89 and 0.88 for cognitive load, and 0.97 and 0.97 for expertise. Our results using the unsupervised autoencoders and random forest classifier achieve an accuracy of 0.99 and an F1 score of 0.98 for arousal classification. 

% Our results, along with the results of the state-of-the-art related work demonstrate the ability for the dataset that we collected to be effectively utilized for a variety of biosignal classification applications. 

Lastly, the comparison between our results and the related works for the MAHNOB-HCI dataset are shown in Table \ref{MAHNOB Comparison}. The best result was found in \cite{siddharth2019} with an accuracy of 0.82 and an F1 score of 0.75 using a uni-modal EDA representation extracted by a CNN with an ELM classifier. Our result outperformed their results with an increase in accuracy and F1 score of 0.01 to 0.83 and 0.76, respectively.

\section{Conclusions and Future Work} \label{conclusions}
% - in thesis: Summary
In this study we examined the use of stacked convolutional autoencoders for learning multi-modal latent representations from wearable biosignal data for subsequent use with a random forest classifier for arousal classification. We utilized 4 datasets and compared our proposed framework with a number of other techniques for feature extraction and machine learning.  Our method showed an overall superior performance with the other methodologies examined with an accuracy and F1 score of 0.89 and 0.91 respectively. Specifically, we obtained accuracies and F1 scores of 0.93 and 0.95 for AMIGOS, 0.79 and 0.86 for ASCERTAIN, 0.99 and 0.98 for CLEAS, and 0.83 and 0.76 for the MAHNOB-HCI datasets. Moreover, our results outperformed the previous state-of-the-art results using the same modalities for arousal classification with the AMIGOS, ASCERTAIN, and MAHNOB-HCI datasets. Arousal detection was performed for the first time on CLEAS. We believe that our method provides a valuable solution for unsupervised learning of multimodal ECG and EDA representations along with supervised learning of affect using a random forest. Our solution performs accurately and is capable of aggregating several datasets to leverage a multi-corpus approach, all the while reducing reliance on human supervision. 

Despite the advantages of our method in reduced reliance on human-annotated data, there are a number of areas which can be considered for future work. For instance, the ability for the solution to generalize to new data can be further explored by changing the method of training and testing validation. The results in this study were obtained by training and validating the framework with 10-fold cross-validation. While most works in this area, including the state-of-the-art references with which we compared our results also used k-fold cross-validation, the use of leave-one-subject-out schemes can be used to better evaluate generalization to unseen data. Moreover, a leave-one-data-out validation scheme can provide further insight into the performance of the method to unseen full datasets. Additionally, more diverse CNN architectures can be explored for comparison with our architecture. Moreover, architectures that exploit both CNNs and recurrent neural networks can provide additional reference points for the performance of our unsupervised representation learning approach.

In the end, while our method achieved great performance for classification of arousal, in future work, additional affective factors, namely valence or dominance, can be explored. In addition, the problem space can be modified from classification to regression with the aim of estimating the affect scores across different dimensions or as the intensity of a particular emotion.

\newpage
\appendices
\section{}
\begin{table}[!htbp]
     
    \caption{Time and frequency domain ECG features.}
    %\begin{tabular}{l|l}
\begin{tabular}{l|l}
\toprule
\textbf{Feature} & 
\textbf{Description } \\
\midrule

         %\Xhline{2\arrayrulewidth}
%         Feature & Description  \\
         %\Xhline{2\arrayrulewidth}
         \textbf{HR}& \makecell[l]{Average heart rate based on number of R peaks in the \\ window}\\
         \textbf{RR\textsubscript{min}}& 
Minimum value of RR interval\\

         \textbf{RR\textsubscript{max}}& 
Maximum value of RR interval\\

         \textbf{RR\textsubscript{diff}}&
 Difference between RR\textsubscript{max} and RR\textsubscript{min}\\

         \textbf{RR\textsubscript{mean}}&
 Mean value of RR interval\\

         \textbf{RR\textsubscript{SD}}& 
Standard deviation of RR interval\\
         
         \textbf{RR\textsubscript{CV}}& 
Coefficient of Variation of RR intervals\\
        \textbf{RMSSD} & \makecell[l]{Root mean squared of successive differences of RR \\ intervals}\\
         \textbf{SDSD} & \makecell[l]{Standard deviation of successive differences of RR \\ intervals}\\

         \textbf{NN50} & 
Number of RR intervals greater than 50 ms\\

         \textbf{PNN50} & 
Percentage of RR intervals greater than 50 ms\\

         \textbf{ULF} &
 Ultra low frequency band (<0.003.)  Hz\\

        \textbf{VLF} &
 Very low frequency band (0.04--0.003) Hz\\

         \textbf{LF} & 
Low frequency band (0.04--0.15)  Hz\\

         \textbf{HF} & 
High frequency band (0.15--0.4) Hz\\

         \textbf{TP} &
 Total power (0--0.4) Hz\\

         \textbf{LF\textsubscript{norm}} & 
Normalized low frequency\\

         \textbf{HF\textsubscript{norm}} &
 Normalized high frequency\\

         \textbf{LF/HF} & 
Ratio of low to high frequency power \\

         \textbf{LMHF} & 
\makecell[l]{Sympatho vagal balance ratio, using mid frequency \\ range of 0.08--0.15 Hz}\\

         %\Xhline{2\arrayrulewidth}
\bottomrule
    \end{tabular}
    \label{ECG_Features}
\end{table}

\begin{table}[!htbp]
    \centering
     
    \caption{Time and frequency domain EDA features.}
  %  \begin{tabular}{l|l}
\begin{tabular}{l|l}
\toprule        
% \Xhline{2\arrayrulewidth}
         \textbf{Feature} & \textbf{Description}  \\
        % \Xhline{2\arrayrulewidth}
\toprule
         \textbf{Num} SCR & Number of SCR events in a given window\\
         \textbf{RT} & Rise time from SCR onset to peak response\\
         \textbf{HRT} & Half recovery time of the SCR peak\\
         \textbf{Amp} & Amplitude of the SCR peak\\
         \textbf{Area} & Area of the SCR peak\\
         \textbf{Prom} & Prominence of the SCR relative to the SCL \\
         \textbf{SCL} & \makecell[l]{Skin conductance level, the average \\ electrodermal response}\\
         \textbf{MAV1Diff SCL} & \makecell[l]{First derivative of the mean absolute value of \\ the SCL}\\
         \textbf{MAV2Diff SCL} & \makecell[l]{Second derivative of the mean absolute value of \\ the SCL}\\
         \textbf{TP} & Total power power of the EDA signal\\
         \textbf{PSD} & \makecell[l]{Power spectrum density estimate of the EDA \\ signal}\\
   %      \Xhline{2\arrayrulewidth}
\bottomrule    
\end{tabular}
    \label{EDA_Features}
\end{table}

% you can choose not to have a title for an appendix
% if you want by leaving the argument blank

% use section* for acknowledgment
\ifCLASSOPTIONcompsoc
  % The Computer Society usually uses the plural form
  \section*{Acknowledgments}
\else
  % regular IEEE prefers the singular form
  \section*{Acknowledgment}
\fi

The authors would like to acknowledge the Canadian Department of National Defense who partially funded this work. I also want to thank Pritam Sarkar, Dr. Dirk Rodenburg, Dr. Aaron Ruberto, Dr. Adam Szulewski, and Dr. Daniel Howes for their collaborations thorough this study.

% Can use something like this to put references on a page
% by themselves when using endfloat and the captionsoff option.
\ifCLASSOPTIONcaptionsoff
  \newpage
\fi

\bibliographystyle{IEEEtran}
% argument is your BibTeX string definitions and bibliography database(s)
\bibliography{main}

\end{document}